\documentclass[times]{TRR}

\usepackage{moreverb,url}
\usepackage{xcolor}
\usepackage[colorlinks,bookmarksopen,bookmarksnumbered,citecolor=red,urlcolor=red]{hyperref}

\newcommand\BibTeX{{\rmfamily B\kern-.05em \textsc{i\kern-.025em b}\kern-.08em
T\kern-.1667em\lower.7ex\hbox{E}\kern-.125emX}}

\def\volumeyear{2020}

\begin{document}

\runninghead{Lee, Lin, and Gran}

\title{DistTune: Distributed Fine-Grained Adaptive Traffic Speed Prediction for Growing Transportation Networks}

\author{Ming-Chang Lee\affilnum{1}, Jia-Chun Lin\affilnum{1}, and Ernst Gunnar Gran\affilnum{1,2}}

\affiliation{\affilnum{1}Department of Information Security and Communication Technology, Norwegian University of Science and Technology, Gjøvik, Norway\\
\affilnum{2}Simula Research Laboratory, Fornebu, Norway\\
Note: This is a draft preprint of a paper to be published in Transportation Research Record. The final paper may be slightly different from this version. Please use the following citation for this paper:
Ming-Chang Lee, Jia-Chun Lin, and Ernst Gunnar Gran, “DistTune: Distributed Fine-Grained Adaptive Traffic Speed Prediction for Growing Transportation Networks,” Transportation Research Record. May 2021. DOI: https://doi.org/10.1177/03611981211011170.
}

\corrauth{Jia-Chun Lin, jia-chun.lin@ntnu.no}

\begin{abstract}
Over the past decade, many approaches have been introduced for traffic speed prediction. However, providing fine-grained, accurate, time-efficient, and adaptive traffic speed prediction for a growing transportation network where the size of the network keeps increasing and new traffic detectors are constantly deployed has not been well studied. To address this issue, this paper presents DistTune based on Long Short-Term Memory (LSTM) and the Nelder-Mead method. Whenever encountering an unprocessed detector, DistTune decides if it should customize an LSTM model for this detector by comparing the detector with other processed detectors in terms of the normalized traffic speed patterns they have observed. If similarity is found, DistTune directly shares an existing LSTM model with this detector to achieve time-efficient processing. Otherwise, DistTune customizes an LSTM model for the detector to achieve fine-grained prediction. To make DistTune even more time-efficient, DistTune performs on a cluster of computing nodes in parallel. To achieve adaptive traffic speed prediction, DistTune also provides LSTM re-customization for detectors that suffer from unsatisfactory prediction accuracy due to for instance traffic speed pattern change. Extensive experiments based on traffic data collected from freeway I5-N in California are conducted to evaluate the performance of DistTune. The results demonstrate that DistTune provides fine-grained, accurate, time-efficient, and adaptive traffic speed prediction for a growing transportation network.
\end{abstract}

\maketitle

\section{Introduction}
Traffic speed is a key indicator to measure the efficiency of a transportation network. Accurate traffic speed prediction is therefore crucial to achieve proactive traffic management and control for transportation networks. During the past decade, many approaches and methods have been introduced for traffic speed prediction. Each of them can be summarized as learning a mapping function between input variables and output variables. These methods can be classified into two main types: parametric and nonparametric. Parametric approaches simplify the mapping function to a known form, i.e., they require a predefined model. A typical example is the autoregressive integrated moving average approach (ARIMA) \cite{1}. On the contrary, nonparametric approaches do not require a predefined model structure. Typical examples include the k-nearest neighbors (k-NN) method \cite{14,15}, artificial neural network (ANN) \cite{32}, and recurrent neural network (RNN) \cite{33}.

However, providing fine-grained, accurate, time-efficient, and adaptive traffic speed prediction for a growing transportation network where the size of the network keeps increasing and new traffic detectors are constantly deployed on the network has not been well studied. To address this issue, this paper proposes a solution based on long short-term memory (LSTM) \cite{2}, which is a special type of RNN. Prior studies such as \cite{3,4,5} have shown that LSTM is superior in time series prediction and provides better prediction accuracy than many existing approaches and neural networks, including Elman NN \cite{3}, Time-delayed NN \cite{3}, Nonlinear Autoregressive NN \cite{3}, support vector machine \cite{3}, ARIMA \cite{1}, and the Kalman Filter approach \cite{34}. Therefore, LSTM is chosen as our building block.

However, several challenges exist and several issues must be addressed in order to achieve the above-mentioned goal (i.e., fine-grained, accurate, time-efficient, and adaptive traffic speed prediction for a growing transportation network). For instance, detectors such as loop sensors and traffic cameras in a transportation network are deployed in different places to collect and monitor traffic data. Depending on the density of nearby population and other factors, the traffic speed observed/collected by detectors at different locations may not be the same. For example, Figure 1 illustrates traffic speed observed by five detectors deployed on freeway I5-N in California \cite{29} in a typical weekday. We can see that the observed traffic speed were similar to each other between 4 a.m. and 6 a.m. However, the pattern of the traffic speed became different and diverse from 6 a.m. Therefore, we recommend that each detector should have its own LSTM model to predict the traffic speed of its coverage so as to provide fine-grained traffic speed prediction and achieve better transportation management and services. However, such an approach is expensive and impractical because training an LSTM model for each individual detector in the growing transportation network is required and that each training process is in general time-consuming. 

\begin{figure*}[!ht]
  \centering
  \includegraphics[width=0.7\textwidth]{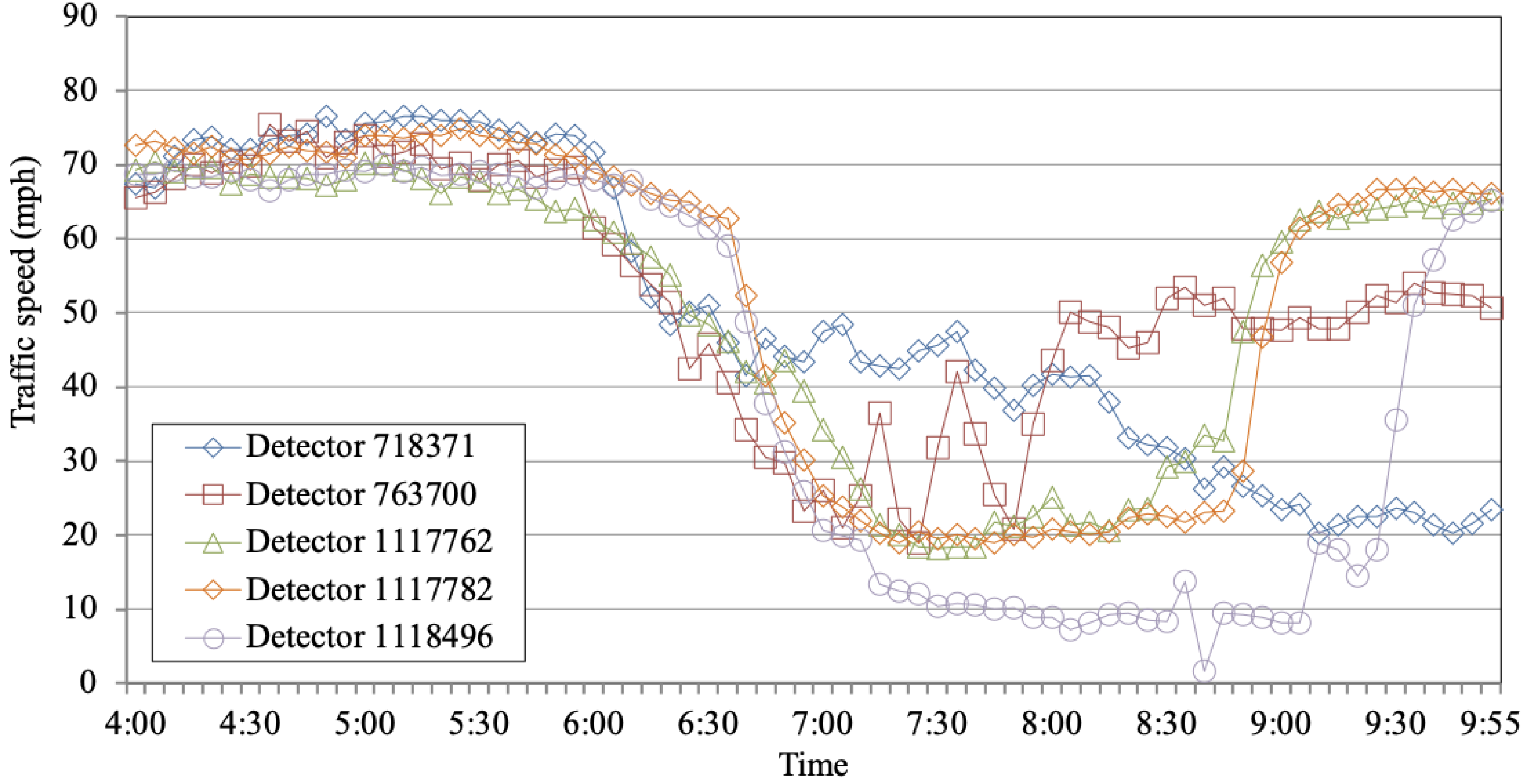}
  \caption{The traffic speed collected by five detectors on freeway I5-N in California between 4 a.m. and 10 a.m. in a typical weekday \cite{36}.}\label{fig:Figure1}
\end{figure*}

Another issue is how to achieve satisfactory prediction accuracy and continuously maintain satisfactory accuracy over time for every single detector in the growing transportation network. It is well-known that the success of LSTM to achieve satisfactory prediction accuracy relies on an appropriately configured set of hyperparameters, which are parameters whose values are set before the training process of LSTM starts. These hyperparameters include learning rate and the number of hidden layers. Determining appropriate values for LSTM hyperparameters is usually done manually by trial and error, which might be time-consuming, and may not even be able to guarantee good prediction accuracy. In addition, an LSTM model might not be able to keep offering satisfactory prediction accuracy because the traffic speed pattern observed by the corresponding detector may change over time.

To summarize, this paper attempts to address the following challenges:
\begin{enumerate}
\item How can we automatically customize an LSTM for each single detector (i.e., appropriately configuring LSTM hyperparameters) in a growing transportation network such that the corresponding LSTM model provides satisfactory prediction accuracy?
\item How can we time-efficiently perform automatic LSTM customization for the increasing number of detectors in growing transportation networks? 
\item How can we keep maintaining satisfactory prediction accuracy for every single detector in a time-efficient way?
\end{enumerate}

To address the above challenges, in this paper, we propose DistTune, which is a distributed scheme to automatically customize LSTM models and constantly provide satisfactory prediction accuracy for every single detector in a growing transportation network. DistTune customizes an LSTM model for a detector by automatically tuning LSTM hyperparameter values and training the corresponding LSTM models based on the Nelder-Mead method (NMM) \cite{6}, which is a commonly applied method used to find the minimum or maximum of an objective function in a multidimensional space. The reason why NMM is employed will be explained in Section 3. 

However, simply using NMM to gradually customize an LSTM model for every single detector is not a scalable solution since the number of detectors in a growing transportation network might be very large and increasing. To enhance scalability and efficiency, DistTune allows detectors to share the same LSTM model if these detectors have observed similar traffic patterns. In addition, DistTune is designed in an incremental, distributed, and parallel manner, and runs on a cluster of computing nodes to accelerate the all required LSTM customization processes. 

Whenever DistTune encounters an unprocessed detector \emph{D\textsubscript{i}}, it checks if the traffic speed pattern observed by \emph{D\textsubscript{i}} is similar to that observed by any other detector whose LSTM models has been customized by DistTune. If the answer is positive and the traffic speed pattern observed by \emph{D\textsubscript{i}} is similar to that observed by detector \emph{D\textsubscript{j}}, DistTune directly shares \emph{D\textsubscript{j}}’s LSTM model with \emph{D\textsubscript{i}} without customizing an LSTM model for \emph{D\textsubscript{i}}. However, if the answer is negative, DistTune requests an available computing node from the cluster to customize an LSTM model for \emph{D\textsubscript{i}}. To guarantee and maintain satisfactory prediction accuracy, DistTune keeps track of the prediction accuracy of all detectors. If any LSTM model is unable to achieve the desired prediction accuracy due to traffic pattern change or other reasons, DistTune requests an available computing node to re-customize an LSTM model for the corresponding detector. 

To demonstrate the performance of DistTune, we conducted three extensive experiments on an Apache Hadoop YARN cluster using the traffic data collected by detectors on freeway I5-N in California. In the first experiment, we designed several scenarios to evaluate the auto-tuning effectiveness of DistTune. The results show that auto-tuning LSTM hyperparameters leads to higher prediction accuracy than manually configuring LSTM hyperparameters. In the second experiment, we evaluated the auto-sharing LSTM performance of DistTune. The results indicate that DistTune significantly reduces the time for LSTM customization. In the last experiment, we evaluated the LSTM re-customization performance of DistTune under two types of re-customization. One is to start from scratch. The other one is based on Transfer Learning \cite{7}. The results surprisingly show that following the concept of Transfer Learning does not bring significant benefits if time consumption and prediction accuracy are both considered. The contributions of this paper are as follows:
\begin{enumerate}
\item Two well-known hyperparameter optimization approaches are evaluated and compared through empirical experiments. One is NMM, and the other is the Bayesian optimization approach \cite{8}. The results suggest that the former method suits DistTune better due to its efficiency in finding appropriate LSTM hyperparameter values for achieving the desired prediction accuracy.
\item The proposed DistTune enables detectors to share the same LSTM model, and it provides LSTM re-customization service for all detectors when any of them fails to provide satisfactory prediction accuracy. The design of DistTune addresses the time-consuming and computation-intensive issues of LSTM customization for the tremendous number of detectors in a growing transportation network. 
\item The performance of DistTune is carefully evaluated through extensive experiments based on traffic data collected by detectors on freeway I5-N in California. The results suggest that DistTune provides fine-grained, accurate, time-efficient, and adaptive traffic speed prediction for a growing transportation network.
\end{enumerate}

The rest of the paper is organized as follows: Section 2 provides background and related work. Section 3 introduces and explains why NMM is chosen by DistTune to autotune LSTM hyperparameters. In Sections 4 and 5, we introduce the details of DistTune and evaluate the performance of DistTune, respectively. Section 6 discusses the training dataset we chose for conducting our experiments. Section 7 concludes this paper and outlines future work. 
\section{BACKGROUND AND RELATED WORK}
In this section, we briefly introduce LSTM and discuss related work.
\subsection{LSTM and its hyperparameters}

LSTM is a special type of RNN. Its architecture is similar to that of RNN except that the nonlinear units in the hidden layers are memory blocks. Each memory block consists of memory cells, an input gate, a forget gate, and an output gate. The input gate decides if the input should be stored in the memory cells. The forget gate determines if current memory contents should be deleted. The output gate decides if current memory contents should be output. This design enables LSTM to preserve information over long time lags, therefore addressing the vanishing gradient problem \cite{9}.

According to \cite{36}, the prediction performance of LSTM greatly depends on configuring appropriate values for the following hyperparameters:
\begin{itemize}
	\item Learning rate
	\item The number of hidden layers
	\item The number of hidden units
	\item Epochs
\end{itemize}

The \textbf{learning rate} controls the amount the weights of LSTM are updated during training. The lower the value, the less is the possibility to miss any local minima, but it might prolong the training process. A  \textbf{hidden layer} is where hidden units take in a set of weighted inputs and produce an output through an activation function. More hidden layers are usually required to learn a large and complex training dataset. A \textbf{hidden unit} is a neuron in a hidden layer. An inappropriate number of hidden units might cause either overfitting or underfitting. An \textbf{epoch} is defined as one forward pass and one backward pass of all the training data. Too few epochs may underfit the training data, whereas too many epochs might overfit the training data.

It is clear that all above-mentioned hyperparameters are important to the learning performance and computation efficiency of LSTM. Therefore, DistTune takes all these hyperparameters into consideration when conducting LSTM customization.

\subsection{Related Work}

Over the past two decades, many traffic prediction approaches have been proposed. As briefly mentioned in the introduction, they can be classified into two categories: parametric approaches and nonparametric approaches. In parametric approaches, a model structure (i.e., the mapping function between input variables and output variables) needs to be determined beforehand based on some theoretical assumptions. Usually the model parameters can be derived from empirical data. The ARIMA model is a typical and widely used parametric approach \cite{10}. Many ARIMA-based approaches were also introduced to improve prediction accuracy, including \cite{11,12,13}.

Unlike parametric approaches, nonparametric approaches do not require a predefined model structure (i.e., there is no need to make assumptions about the mapping function). Typical examples include k-NN, ANN, RNN, hybrid approaches, etc. The k-NN method was used in \cite{14} to forecast freeway traffic. Several variants of the k-NN and RNN methods were then introduced for traffic prediction, such as \cite{15} and \cite{37}. Le et al. \cite{16} addressed traffic speed prediction using big traffic data obtained from static sensors and proposed local Gaussian Processes to learn and make predictions for correlated subsets of data. Jiang and Fei \cite{17} introduced a data-driven vehicle speed prediction method based on Hidden Markov models. Ma et al. \cite{3} employed LSTM to forecast traffic speed using remote microwave sensor data. However, all the above approaches focus on predicting traffic on a fix-length road section or a fix-sized region. Unlike these approaches, DistTune is a flexible scheme for a growing transportation network 
since it is able to incrementally customize LSTM models for any detectors that are newly deployed on a transportation network. This property makes DistTune a flexible and scalable solution for a growing transportation network.

Ma et al. \cite{18} used deep learning theory to predict traffic congestion evolution in large-scale transportation networks. Furthermore, in \cite{19}, Ma et al. predicted traffic speed in large-scale transportation networks by representing network traffic as images and employing convolutional neural networks to make predictions. However, both of these methods require the scale of the target transportation network to be fixed and specified in advance, which is not required when DistTune is employed. DistTune can handle an increasing number of detectors on the fly without requiring specifying the scale of the target transportation networks beforehand.

More recently, Lee et al. \cite{35} formulated the problem of customizing an LSTM model into a finite Markov decision process and then introduced a distributed approach called DALC to automatically customize LSTM models for detectors in large-scale transportation networks. However, DALC only focuses on two LSTM hyperparameters (i.e., the number of hidden layers and epochs) using a fixed state transition graph. Different from DALC, our DistTune considers two more LSTM hyperparameters without using a finite Markov decision process. Hence, DistTune is more flexible than DALC.

In year 2020, Lee et al. \cite{36} introduced DistPre to provide find-grained traffic speed prediction for large-scale transportation networks based on LSTM customization and distributed computing. However, the LSTM models generated by DistPre are unable to be updated to adapt to traffic pattern changes. Hence, DistPre might not be able to maintain satisfactory prediction accuracy for every single detector in a growing transportation network over time.

\section{EVALUATION OF BOA AND NMM}
Hyperparameter optimization is the process of finding appropriate values for the hyperparameters of a training algorithm such that the algorithm can achieve desirable results. Existing approaches include NMM, Bayesian optimization approach (BOA), grid search, random walk, random search, genetic algorithm, greedy search, simulated annealing, particle swarm optimization, etc. The evaluation of these approaches can be found in, for examples, \cite{20} and \cite{21}. 

Among these approaches, BOA has gained great popularity in recent years in a wide range of areas due to its power and efficiency. BOA was designed to find the optimal value of a black-box objective function. In our context, the black-box objective function refers to an LSTM model, and the optimal value refers to a hyperparameter setting (i.e., assigning a value to each LSTM hyperparameter we consider). With this optimal hyperparameter setting, the LSTM model is able to provide satisfactory prediction accuracy. 

In BOA, the uncertainty of the objective function across not-yet evaluated values is modeled as a prior probability distribution (called prior), which captures our beliefs about the behavior of the function. After gathering the function evaluations (i.e., the prediction accuracy of an LSTM model under a particular hyperparameter setting), the prior is updated to form the posterior distribution over the objective function. The posterior distribution is then used to construct an acquisition function that will select the most promising value (i.e., the most promising hyperparameter setting) for next evaluation. The above process iterates toward an optimum. Examples of acquisition functions include, for instances, probability of improvement, expected improvement, and Bayesian expected losses \cite{22}. All of them try to use and balance exploration and exploitation to minimize the number of function queries. This is why BOA is suitable for functions that are expensive to evaluate. An in-depth review of BOA can be found in \cite{23}.

NMM is another popular optimization method for non-linear functions. This method does not require any derivative information, making it suitable for problems with non-smooth functions. NMM minimizes an objective function by generating an initial simplex based on a predefined vertex and then performing a function evaluation at each vertex of the simplex. Note that a simplex has $\emph{n}+1$ vertices in R\emph{\textsuperscript{n}} where  \emph{n} is the number of dimensions of the parameter space. For instance, the simplex is a triangle when  \emph{n} is 2. In our context, a vertex is a set of values assigned to LSTM hyperparameters, the predefined vertex is a default LSTM hyperparameter setting, and the function evaluation is to derive the prediction error of an LSTM trained with a certain dataset under a hyperparameter setting. 

A sequence of transformations is then performed iteratively on the simplex, aiming to decrease the function values at its vertices. Possible transformations include reflection, expansion, contraction, and shrinking. The above process terminates when the sample standard deviation of the function values of the current simplex fall below some tolerance. We refer readers to \cite{24} for more details about these transformations.

In this paper, we focus on evaluating BOA and NMM to see which of them suits DistTune better, i.e., which of them can more time efficiently find LSTM hyperparameters for detectors such that the resulting LSTM models are able to achieve the desired prediction accuracy. In this experiment, BOA and NMM focus on auto-tuning the four abovementioned LSTM hyperparameters, i.e., learning rate (denoted by \emph{R\textsubscript{Learn}}), the number of hidden layers (denoted by \emph{N\textsubscript{Layer}}), the number of hidden units (denoted by \emph{N\textsubscript{Unit}}), and the number of epochs (denoted by \emph{ep}). Table~\ref{Table1} presents the domains of these hyperparameters. 
\begin{table*}[!ht]
	\caption{Four Hyperparameters and their Domains}\label{Table1}
	\begin{center}
		\begin{tabular}{l||l l l l}
		\hline
			Hyperparameter & \emph{R\textsubscript{Learn}} & \emph{N\textsubscript{Layer}} & \emph{N\textsubscript{Unit}} & \emph{ep} \\\hline\hline
			Domain & [0.01, 0.2] & [1, 10] & [2, 40] & [100, 1000] \\\hline
			Discrete with step   & 0.01 & 1 & 2 & 20 \\\hline
		\end{tabular}
	\end{center}
\end{table*}

To fairly evaluate and compare BOA and NMM, we selected the five detectors shown in Figure 1 to be their targets. It is clear from Figure 1 that the traffic speed patterns observed by detectors 1117762 and 1117782 are similar to each other, and the traffic speed patterns observed by the other three detectors are diverse. These phenomena reflect the traffic speed patterns observed by detectors in a real transportation network. This is why we chose these five detectors to evaluate BOA and NMM. If one of these two approaches is able to more time-efficiently auto-tune LSTM hyperparameters for these five detectors than the other such that the corresponding models achieve the desired prediction accuracy, it is likely that this approach can offer better time-efficiency than the other when it is adopted by DistTune in a growing transportation network.

In this evaluation, each of these five detectors has the same size of training data (i.e., traffic speed values from 5 working days with the collection interval of every 5 minutes). Two metrics were used for the comparison: average absolute relative error (AARE) and LSTM customization time (LCT). AARE is a well-known measure for the prediction accuracy of a forecast method \cite{30}, which is defined as follows:

\begin{equation}
\emph{\emph{AARE}}=\frac{1}{W} \sum_{w=1}^{W} \frac{\mid{s\textsubscript{\emph{w}}-\widehat{s\textsubscript{\emph{w}}}\mid}}{s\textsubscript{\emph{w}}}
\end{equation} 
where \emph{W} is the total number of data points considered for comparison, \emph{w} is the index of data point, \emph{s\textsubscript{w}} is the actual traffic speed value at \emph{w}, and $\widehat{\emph{s\textsubscript{w}}}$ is the forecast traffic speed value at \emph{w}. The lower AARE is, the higher prediction accuracy the forecast method has. The second metric (i.e., LCT) is the time taken by BOA and NMM to individually find a LSTM hyperparameter setting for a detector such that the prediction accuracy of the corresponding LSTM model satisfies a predefined AARE threshold, which is 0.05 in this paper (This value is considered satisfactory according to \cite{31}, \cite{35}, and \cite{36}).

In this experiment, both approaches start with the following default hyperparameter setting, taken from \cite{36}:
\begin{center}
$<$\emph{R\textsubscript{Learn}}=0.01, \emph{N\textsubscript{Layer}}=1, \emph{N\textsubscript{Unit}}=2, \emph{ep}=100$>$
\end{center}
As soon as they find a hyperparameter setting with which the corresponding LSTM model has an AARE less than or equal to the AARE threshold, they automatically terminate. However, if they have iterated for 20 times without being able to find such a hyperparameter setting, they will be forcibly terminated. This experiment was performed on a laptop running MacOS 10.13.1 with 2.5 GHz Quad-Core Intel Core i7 and 16GB 1600 MHz DDR3.

Figure 2 illustrates the performance of BOA and NMM in terms of AARE (presented as marked lines) and LCT (presented as clustered columns). We can see that all the LSTMs customized by NMM satisfy the AARE threshold because their AARE values are all lower than 0.05. However, not all the LSTMs customized by BOA achieve the same good result, especially for detector 1118496. After BOA iterates 20 times, the best AARE value for this detector is still higher than 0.05, and the LCT taken by BOA has reached 1006 minutes. 
\begin{figure*}[!ht]
  \centering
  \includegraphics[width=0.7\textwidth]{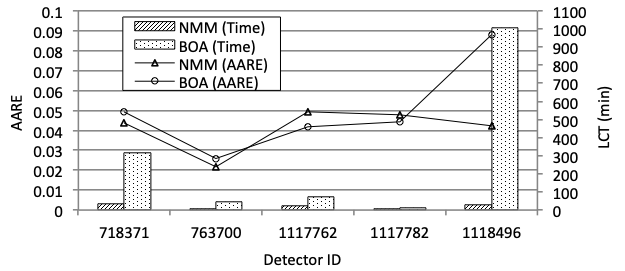}
  \caption{The performance of NMM and BOA on customizing LSTM models for five detectors that observe different traffic speed patterns. AARE stands for average absolute relative error, and LCT stands for LSTM customization time.}\label{fig:Figure2}
\end{figure*}

Based on the above results, we conclude that NMM is more suitable than BOA for DistTune because NMM has better efficiency and effectiveness in finding appropriate values for LSTM hyperparameters such that the desired prediction accuracy for detectors can be satisfied. Even though BOA is able to find optimal hyperparameter settings that might be able to reach higher prediction accuracy than NMM, its high time consumption is unaffordable for DistTune. This is why NMM is adopted by DistTune to autotune LSTM hyperparameters.

\section{Methodology of DistTune}
In this section, we firstly introduce all algorithms of DistTune and then suggest a customization protocol for practitioners to implement DistTune on their growing transportation networks.
\subsection{Algorithms}
DistTune is designed and implemented as an incremental LSTM auto-tuning, sharing, and re-customization system on a cluster consisting of a master server and a set of worker nodes. The master server determines whether to customize, share, or re-customize LSTM models for individual detectors. Each worker node waits for an instruction from the master server to conduct LSTM customization/re-customization for a given detector.

Figure 3 illustrates the LSTM auto-tuning and sharing algorithm running on the master server, and Figure 4 shows the high-level flowchart of the algorithm. Let \emph{G=D\textsubscript{1},D\textsubscript{2}, ..., D\textsubscript{x}} be a list of detectors that have their own LSTM models customized by DistTune. Note that \emph{G} is empty before DistTune is employed and launched. Whenever encountering an unprocessed detector (denoted by \emph{U\textsubscript{i}}), the algorithm first normalizes a list of traffic speed values observed and collected by \emph{U\textsubscript{i}}. Let \emph{L\textsubscript{i}} be the list, and \emph{L\textsubscript{i}}=\{\emph{v\textsubscript{i,1},v\textsubscript{i,2}, ..., v\textsubscript{i,T}}\}  where \emph{v\textsubscript{i,t}} is the traffic speed value collected by \emph{U\textsubscript{i}} at time \emph{t, t=1,2,…,T}. The normalization process divides \emph{v\textsubscript{i,t}} by \emph{V}, which is a predefined fixed value (for example, 70 to represent speed limit in mph). Let  \emph{Nor\textsubscript{i}} be the normalization result, i.e., \emph{Nor\textsubscript{i}}=\{\emph{n\textsubscript{i,1}, n\textsubscript{i,2}, ..., n\textsubscript{i,T}}\} where  $\emph{n\textsubscript{i,T}}=\frac{\emph{v\textsubscript{i,T}}}{V}$. 

The master server decides if it should customize an LSTM model for \emph{U\textsubscript{i}} by sequentially comparing \emph{U\textsubscript{i}} with every detector (denoted by \emph{D\textsubscript{j}}, \emph{j=1,2,…,x}) in \emph{G} in terms of their normalized traffic speed pattern based on the following equation \cite{36}: 
\begin{equation}
\emph{\emph{AARD}}\textsubscript{\emph{i.j}}=\frac{1}{T} \sum_{i=1}^{T} \frac{\mid{n\textsubscript{\emph{i,t}}-n\textsubscript{\emph{j,t}}}\mid}{n\textsubscript{\emph{i,t}}}
\end{equation}
where \emph{\emph{AARD}}\textsubscript{\emph{i.j}} is the average absolute relative difference between the traffic speed patterns collected by \emph{U\textsubscript{i}} and \emph{D\textsubscript{j}}, and \emph{n\textsubscript{j,t}} is the normalized traffic speed value collected by \emph{D\textsubscript{j}} at time \emph{t}, where $\emph{n\textsubscript{j,t}}=\frac{\emph{v\textsubscript{j,t}}}{V}$.

In fact, Equation 2 is similar to the AARE equation shown in Equation 1. Recall that a smaller AARE value indicates that what the prediction method predicts is more similar to the actual one. We follow the same concept to propose Equation 2 and use it to measure if two detectors have observed similar normalized traffic speed patterns. If \emph{\emph{AARD}}\textsubscript{\emph{i.j}} is less than a predefined threshold \emph{thd}\textsubscript{\emph{AARD}} (implying that \emph{U\textsubscript{i}} and \emph{D\textsubscript{j}} observe a similar normalized traffic speed pattern), the master server directly shares \emph{D\textsubscript{j}}’s LSTM model with \emph{U\textsubscript{i}}, without customizing an LSTM model for \emph{U\textsubscript{i}} (see Figure 3: line 9 to line 12).

However, if the master server is unable to find any detector that has observed a similar normalized traffic speed pattern with \emph{U\textsubscript{i}} (i.e., line 13 holds), the master server asks an available worker node to customize an LSTM model for \emph{U\textsubscript{i}}, and then appends \emph{U\textsubscript{i}} to the end of \emph{G} to indicate that \emph{U\textsubscript{i}} has its own customized LSTM model. Based on how each detector is appended to \emph{G}, it is clear that every detector in \emph{G} must have observed a distinct traffic speed pattern.

\begin{figure*}[!ht]
  \centering
  \includegraphics[width=0.6\textwidth]{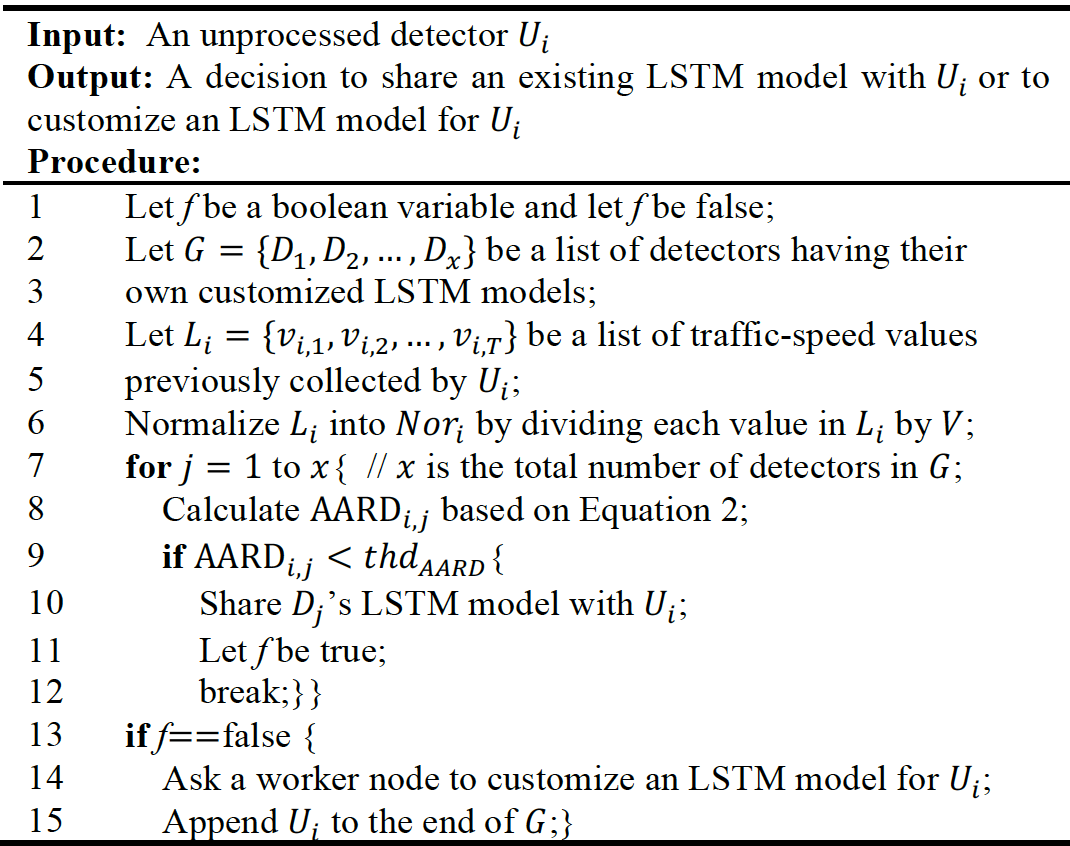}
  \caption{The LSTM auto-tuning and sharing algorithm performed by the master server. }\label{fig:Algorithm1}
\end{figure*}
\begin{figure}[!ht]
  \centering
  \includegraphics[width=0.48\textwidth]{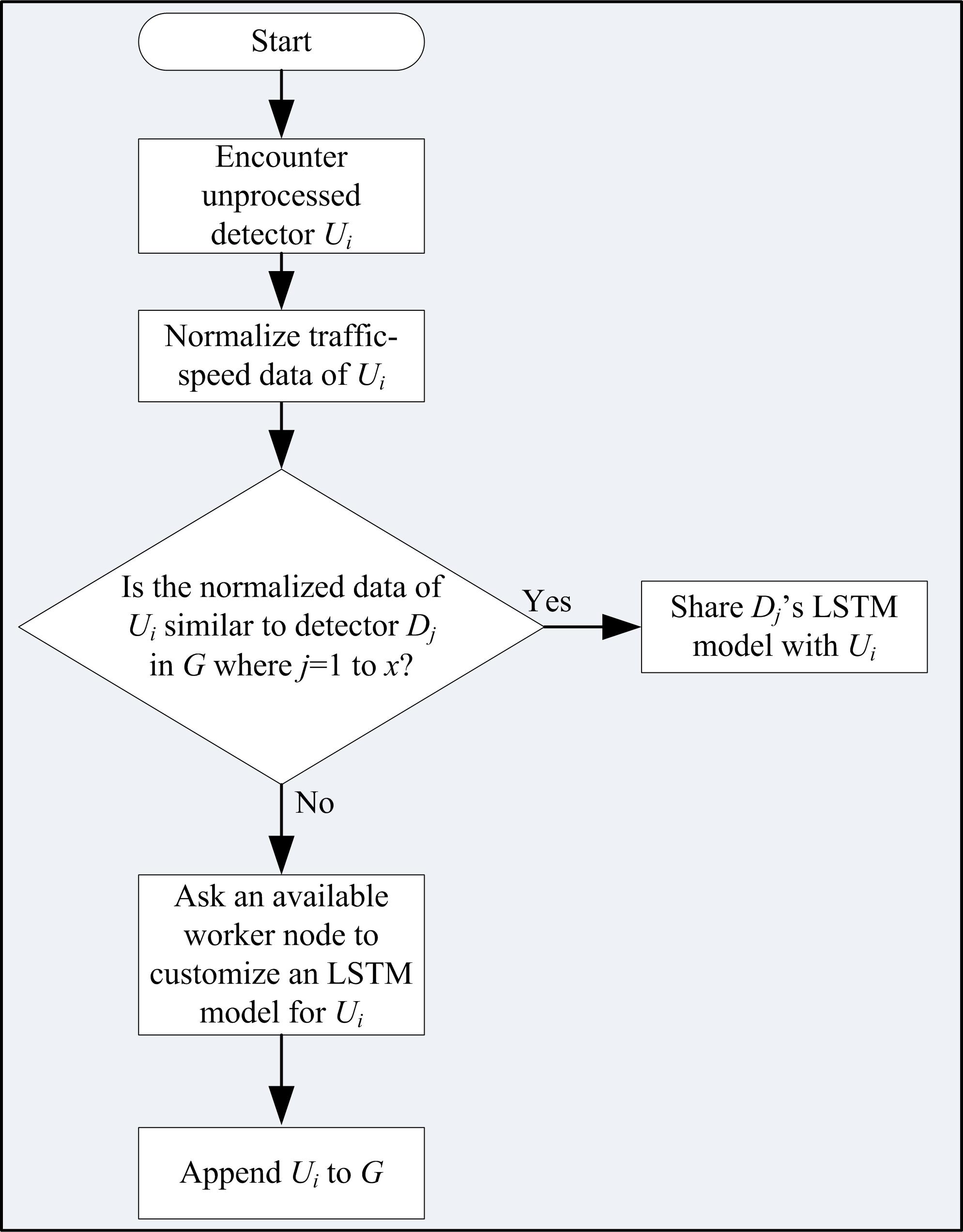}
  \caption{The flowchart of the LSTM auto-tuning and sharing algorithm.}\label{fig:flowchat2}
\end{figure}

Several factors might stop an LSTM model from providing satisfactory prediction accuracy. For example, the traffic pattern collected by a detector has changed and no longer follows the previous pattern that is used to train the detector’s LSTM model. In order to guarantee fine-grained and satisfactory prediction accuracy for every detector in a growing transportation network, the master server keeps track of every single detector by using the tracking algorithm shown in Figure 5.

Let \emph{A} be a list of detectors that have been processed by DistTune, and \emph{A=D\textsubscript{1},D\textsubscript{2}, ..., D\textsubscript{y}} where \emph{y $\geq$ $x$}. In other words, each detector in \emph{A} has an LSTM model either inherited from another detector or customized by DistTune. Periodically, the master server checks if the prediction accuracy of every detector in \emph{A} is still satisfactory. If a detector (denoted by \emph{D\textsubscript{k}}, \emph{k=1,2,…,y} has an AARE value higher than a predefined threshold \emph{thd\textsubscript{AARE}}, the master server requests a worker node to re-customize an LSTM model for \emph{D\textsubscript{k}}. After that, the master server appends \emph{D\textsubscript{k}} to the end of \emph{G} if this is the first time that \emph{D\textsubscript{k}} gets its own LSTM model.

\begin{figure*}[!ht]
  \centering
  \includegraphics[width=0.6\textwidth]{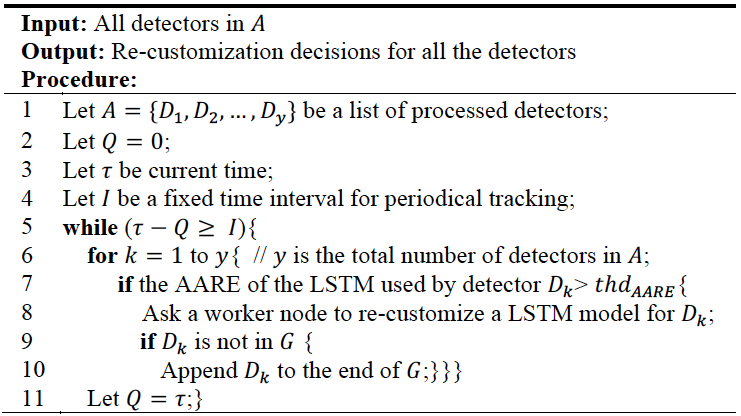}
  \caption{The tracking algorithm performed by the master server.}\label{fig:Algorithm2}
\end{figure*}

Whenever a worker node receives a customization or re-customization request for detector \emph{D\textsubscript{$\varphi$}} (where \emph{D\textsubscript{$\varphi$}}$\in$ A) from the master server, the worker node conducts Figure 6. First, the worker node employs NMM to create the initial simplex based on the default LSTM hyperparameter setting (i.e., \emph{R\textsubscript{Learn}}=0.01, \emph{N\textsubscript{Layer}}=1, \emph{N\textsubscript{Unit}}=2, \emph{ep}=100) since this setting introduces less computational cost \cite{36}. With this setting, the worker node trains a LSTM model with the traffic speed data previously collected by \emph{D\textsubscript{$\varphi$}}. If the AARE of this LSTM model is less than or equal to \emph{thd\textsubscript{AARE}}, the worker node stops NMM and outputs this LSTM because the desired prediction accuracy has been already reached.

\begin{figure*}[!ht]
  \centering
  \includegraphics[width=0.6\textwidth]{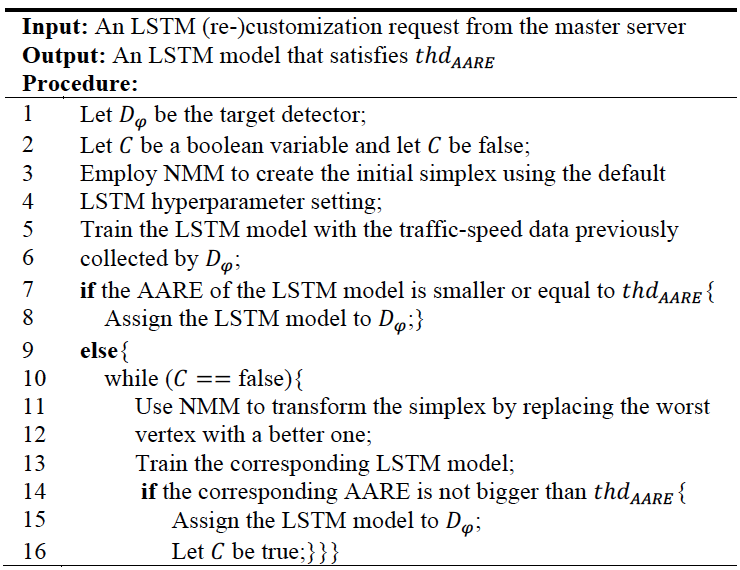}
  \caption{The LSTM (re-)customization algorithm performed by every worker node.}\label{fig:Algorithm3}
\end{figure*}

However, if the AARE is higher than \emph{thd\textsubscript{AARE}}, the worker node keeps using NMM to transform the simplex by replacing the worst vertex with a better one and repeats the same procedure until the function evaluation at a vertex of the simplex is satisfactory. In other words, the worker node terminates when it finds an LSTM model of which the AARE is less than or equal to \emph{thd\textsubscript{AARE}}. The corresponding LSTM model is the final output, and it will be used to predict future traffic speed collected by \emph{D\textsubscript{$\varphi$}}. 

\subsection{Customization protocol}
In order to apply DistTune to a growing transportation network, the following steps should be conducted beforehand:

\begin{enumerate}
\item
Choose a target transportation network.

\item
Collect the traffic speed data observed by all detectors in the target transportation network, and then determine the sizes of training dataset and testing dataset for LSTM customization.

\item
Determine \emph{thd\textsubscript{AARD}}. If the value of \emph{thd\textsubscript{AARD}} is low, it means that detectors need to have high similarity in their normalized traffic speed patterns in order to be able to share a same LSTM model.

\item
Determine \emph{thd\textsubscript{AARE}}. The lower the value of \emph{thd\textsubscript{AARE}} is set, the higher prediction accuracy will be achieved. 

\end{enumerate}

Once one has finished the above steps, DistTune can be launched to initiate its traffic speed prediction service.

\section{PERFORMANCE EVALUATION}
To evaluate DistTune, we conducted three extensive experiments using traffic data from the Caltrans performance measurement system \cite{25}, which is a database of traffic data collected by detectors placed on state highways throughout California. The freeway I5-N was chosen, which is a major route from the Mexico-United States border to Oregon with a total length of 796.432 miles. In these experiments, DistTune incrementally provides its service until 110 detectors on I5-N are covered to show that DistTune is able to be employed in a growing transportation network and to incrementally customize LSTM models for any detectors that are newly deployed to the network. Note that the distance between two consecutive detectors of the 110 detectors is around 5 miles, and each detector collected traffic data every five minutes.
For each detector, we crawled its traffic data for six continuous working days, and then split it into a training dataset (the first 5 days) and a testing dataset (the last day). In other words, the training dataset and testing dataset for each detector consist of 1440 and 288 data points, respectively. Since all the traffic data was aggregated at five-minute intervals, DistTune follows the same interval for prediction. The reason why the training dataset is five continuous working days will be discussed in the next section.

In all the experiments, \emph{thd\textsubscript{AARE}}=0.05 and \emph{thd\textsubscript{AARD}}=0.1 \cite{36}. In this paper, if a detector is able to provide 95\% prediction accuracy, we consider it satisfactory. This is why we set \emph{thd\textsubscript{AARE}} to be 0.05. The same reason for \emph{thd\textsubscript{AARD}}: if two detectors have 90\% similarity in their normalized traffic speed patterns, we consider these patterns similar. This is why we set \emph{thd\textsubscript{AARD}} to be 0.1. In fact, these two thresholds are configurable if one wants to change the degree of similarity or achieve a different level of prediction accuracy.

Three widely used performance metrics \cite{30}, average absolute error (AAE), AARE, and root mean square error (RMSE) were employed in all our experiments to evaluate LSTM prediction accuracy. Please see Equation 1 for AARE. The equations to calculate AAE and RMSE are as follows:

\begin{equation}
\emph{\emph{AAE}}=\frac{1}{X} \sum_{q=1}^{X}\mid{s\textsubscript{\emph{q}}-\widehat{s\textsubscript{\emph{q}}}\mid} 
\end{equation}

\begin{equation}
\emph{\emph{RMSE}}=\sqrt{\frac{1}{X} \sum_{q=1}^{X}{(s\textsubscript{\emph{q}}-\widehat{s\textsubscript{\emph{q}}}})^2}
\end{equation} 
where \emph{X} is the total number of data points for comparison, \emph{q} is the index of a data point, \emph{s\textsubscript{q}} is the actual traffic speed value at \emph{q}, and $\widehat{\emph{s\textsubscript{q}}}$ is the forecast traffic speed value at \emph{q}. Low values for AAE, AARE, and RMSE indicate that the corresponding LSTM model has high prediction accuracy. 

\subsection{Experiment 1}
In this experiment, we designed five scenarios (see Table 2) to evaluate the auto-tuning effectiveness of DistTune by temporarily disabling the LSTM sharing function. In other words, every detector in this experiment will always get its own LSTM model customized by DistTune. In Scenario 1, DistTune only autotunes two hyperparameters. The other two hyperparameters are assigned with small values. In Scenario 2, DistTune autotunes one more hyperparameter. In Scenario 3, all of the four hyperparameters are automatically tuned by DistTune. Scenario 4 and Scenario 5 are similar to Scenario 1 and Scenario 2, respectively, but with an equal or a higher value for each non-autotuned hyperparameters. The goal is to see the impact of these increased values on DistTune.

\begin{table}[!ht]
	\caption{The Hyperparameter Settings in Five Scenarios}\label{Table2}
	\begin{center}
		\begin{tabular}{l|| l l l l l}
		\hline
			Scenario & 1 & 2 & 3 & 4 & 5 \\\hline\hline
			\emph{R\textsubscript{Learn}} & tune & tune & tune & tune & tune \\\hline
			\emph{N\textsubscript{Layer}} & 1 & tune & tune & 2 & tune \\\hline
			\emph{N\textsubscript{Unit}}  & 10 & 6 & tune & 10 & 10 \\\hline
			\emph{ep} & tune & tune & tune & tune & tune \\\hline
		\end{tabular}
	\end{center}
\end{table}

Four performance metrics were used in this experiment: 
\begin{itemize}
	\item Cumulative LSTM customization time: The summation of the time taken by DistTune to customize an LSTM for each detector such that the corresponding LSTM satisfies \emph{thd\textsubscript{AARE}}.
	\item Average AAE: The average AAE of all the LSTMs customized by DistTune. See Equation 5.
	\item Average AARE: The average AARE of all the LSTMs customized by DistTune. See Equation 6
	\item Average RMSE: The average RMSE of all the LSTMs customized by DistTune. See Equation 7.
\end{itemize}

\begin{equation}
\emph{\emph{Average\ AAE}}=\frac{\sum_{r=1}^{Z} {\emph{\emph{AAE}}\textsubscript{\emph{r}}}}{Z} 
\end{equation}
\begin{equation}
\emph{\emph{Average\ AARE}}=\frac{\sum_{r=1}^{Z} {\emph{\emph{AARE}}\textsubscript{\emph{r}}}}{Z} 
\end{equation}
\begin{equation}
\emph{\emph{Average\ RMSE}}=\frac{\sum_{r=1}^{Z} {\emph{\emph{RMSE}}\textsubscript{\emph{r}}}}{Z} 
\end{equation}

In the above three equations, \emph{Z} is the total number of LSTMs (in this experiment, \emph{Z} equals to 110), and \emph{r} is the index number of an LSTM, \emph{r} = 1,2,...,\emph{Z}.

Figure 7 illustrates the cumulative LSTM customization time taken by DistTune in the five scenarios. From the results of the first three scenarios, it seems that the cumulative LSTM customization time increases when more hyperparameters are automatically tuned by DistTune. However, this is not true when the last two scenarios are further considered. We can see that Scenario 4 and Scenario 5 cost more time than Scenario 3, even though they have less autotuned hyperparameters than Scenario 3. In other words, increasing the number of autotuned LSTM hyperparameters does not mean that the corresponding LSTM customization time will always increase. The reason why Scenario 4 resulted in the longest LSTM customization time is that the network structure of each LSTM model in Scenario 4 is more complicated than that in all the other Scenarios. Therefore, using DistTune to autotune \emph{N\textsubscript{Layer}} and \emph{N\textsubscript{Unit}} would be more appropriate.

\begin{figure}[!ht]
  \centering
  \includegraphics[width=0.45\textwidth]{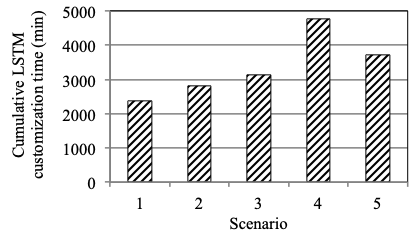}
  \caption{The cumulative LSTM customization time consumed by DistTune in five scenarios when the LSTM sharing function is disabled.}\label{fig:Figure7}
\end{figure}
\begin{figure}[!ht]
  \centering
  \includegraphics[width=0.45\textwidth]{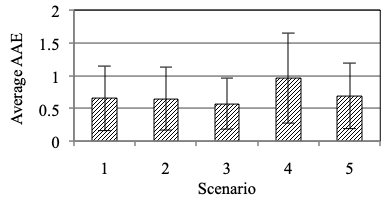}
  \caption{The average AAE results in five scenarios when the LSTM sharing function of DistTune is disabled.}\label{fig:Figure8}
\end{figure}
\begin{figure}[!ht]
  \centering
  \includegraphics[width=0.45\textwidth]{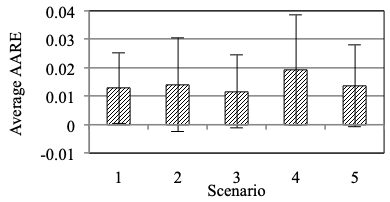}
  \caption{The average AARE results in five scenarios when the LSTM sharing function of DistTune is disabled.}\label{fig:Figure9}
\end{figure}
\begin{figure}[!ht]
  \centering
  \includegraphics[width=0.45\textwidth]{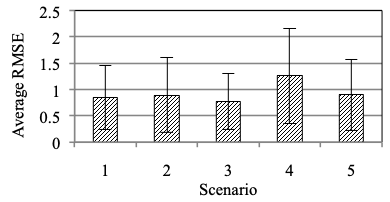}
  \caption{The average RMSE results in five scenarios when the LSTM sharing function of DistTune is disabled.}\label{fig:Figure9}
\end{figure}
Figures 8, 9, and 10 show the prediction performance of DistTune in the five scenarios when the LSTM sharing function is disabled. It is clear that Scenario 3 has the highest prediction accuracy since its average AAE value, average AARE value, average RMSE value, and the corresponding standard deviation values are the smallest among the five scenarios. The results confirm that auto-tuning all the considered hyperparameters leads to the best prediction accuracy. However, if the cumulative LSTM customization time shown in Figure 7 is further considered, it appears that Scenario 3 is more computationally expensive than Scenario 1 and Scenario 2 due to its significant increase in computation time, compared with its slight improvement in prediction accuracy. This is because DistTune in this experiment disables the LSTM sharing function. It needs to individually customize an LSTM model for every single detector, so the cumulative LSTM customization time of DistTune cannot have a significantly improvement. In the next experiment, we will show how DistTune mitigates this issue by enabling the LSTM sharing function.

\subsection{Experiment 2}
In this experiment, we studied the impact of the LSTM sharing function and the number of worker nodes on the performance of DistTune. To this aim, we designed four cases as listed in Table~\ref{Table3}. In both Case 1 and Case 2, the cluster running DistTune has a single worker node. However, the LSTM sharing function of DistTune was disabled in Case 1, whereas it was enabled in Case 2. In Case 3 and Case 4, the cluster has 30 worker nodes, and the sharing function was disabled and enabled, respectively. Two performance metrics were employed in this experiment: total LSTM customization time and average AARE. Note that the former is the total elapsed time from DistTune is launched until all the 110 detectors have obtained LSTM models. 

\begin{table*}[!ht]
	\caption{The Details of Four Cases}\label{Table3}
	\begin{center}
		\begin{tabular}{l|| l l l l}
		\hline
			Case No. & 1 & 2 & 3 & 4 \\\hline\hline
			Number of worker nodes &  \centering 1 & 1 & 30 & 30 \\\hline
			The LSTM sharing function & Disabled & Enabled & Disabled & Enabled \\\hline
		\end{tabular}
	\end{center}
\end{table*}

\begin{figure*}[!ht]
  \centering
  \includegraphics[width=0.7\textwidth]{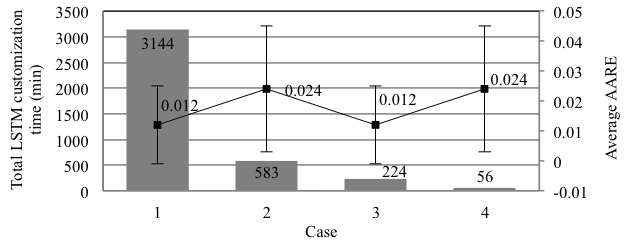}
  \caption{The performance of DistTune in four cases. Note that TLCD represents total LSTM customization time.}\label{fig:Figure11}
\end{figure*}
Figure 11 shows the performance results of DistTune in the four cases. Case 1 is the most time-consuming among the four cases. This is because only one worker node was employed to individually and sequentially customize LSTM models for each of the 110 detectors. This situation turns better in Case 2 since the sharing function was enabled. The total LSTM customization duration reduces by 81.46\%$=\frac{3144-583}{3144}$ from Case 1 to Case 2, implying that allowing LSTM models to be shared between detectors significantly reduces the number of LSTM models that DistTune needs to customize, even though there is only one worker node in the cluster supporting DistTune. 

When the 30 worker nodes were used and the sharing function was disabled, i.e., Case 3, the total LSTM customization time reduces to 224 minutes, meaning that increasing the scale of the cluster and utilizing parallel computing also help improve the performance of DistTune. By further enabling detectors to share their LSTM models without individual customization, i.e., Case 4, the total time drops to only 56 minutes. The reduction is around 75\%$=\frac{224-56}{224}$ compared with Case 3. This significant performance improvement is mainly due to two factors. Firstly, by enabling the LSTM sharing function, DistTune only needed to customize 31 LSTM models for the 110 detectors. Secondly, the task of LSTM customization was shared by the 30 worker nodes and performed in parallel.

On the other hand, from the perspective of average AARE, both Case 1 and Case 3 lead to the same result (around 0.012 as shown in Figure 11). This is because the algorithm used by DistTune to customize LSTM models (i.e., NMM) is deterministic, i.e., the result generated by NMM for a given detector is always identical no matter which worker node performs the task. 

Due to the same reason, the average AARE results in Case 2 and Case 4 are identical, but they are a little higher than those in Case 1 and Case 3. The reason is that not all the detectors in Case 2 and Case 4 have their own customized LSTMs. 

In Case 1 and Case 3, all the 110 detectors require no re-customization since each of them has a customized LSTM model. However, in Case 2 and Case 4, seven out of the 110 detectors require re-customization because the LSTMs that they inherited from other detectors are unable to satisfy \emph{thd\textsubscript{AARE}}. Nevertheless, the re-customization ratio is low (6.36\%$=\frac{7}{110}$), suggesting that setting \emph{thd\textsubscript{AARD}} as 0.1 seems appropriate.

Based on all the above results, we conclude that DistTune in Case 4 provides the best trade-off between time efficiency and prediction accuracy, implying that enabling LSTM sharing and running DistTune on a large cluster are both important for DistTune. 

\subsection{Experiment 3}
In this experiment, we evaluated the LSTM re-customization performance of DistTune by re-customizing LSTM models for the seven detectors that require re-customization in Cases 2 and 4 in Experiment 2. Two types of LSTM re-customization were considered:
\begin{itemize}
	\item Type 1: Based on the default hyperparameter setting.
	\item Type 2: Based on Transfer Learning.
\end{itemize}

In Type 1, DistTune re-customizes an LSTM model for each of the seven detectors by using the default hyperparameter setting (i.e., \emph{R\textsubscript{Learn}}=0.01, \emph{N\textsubscript{Layer}}=1, \emph{N\textsubscript{Unit}}=2, \emph{ep}=100) to generate the initial simplex for NMM. In Type 2, DistTune re-customizes an LSTM model for detector \emph{i} by using the LSTM hyperparameter setting of detector \emph{j} to generate the initial simplex if the LSTM model currently used by detector \emph{i} is shared by detector \emph{j}. 

\begin{table}[!ht]
	\caption{The Hyperparameter Setting in Type 1 and Type 2}\label{Table4}
	\begin{center}
		\begin{tabular}{l|| l l l l}
		\hline
			Re-customization Type & \emph{R\textsubscript{Learn}} & \emph{N\textsubscript{Layer}} & \emph{N\textsubscript{Unit}} & \emph{ep} \\\hline\hline
			Type 1&0.01&1&2&100 \\\hline
			Type 2&0.05&1&10&180\\\hline
		\end{tabular}
	\end{center}
\end{table}
\begin{figure*}[!ht]
  \centering
  \includegraphics[width=0.6\textwidth]{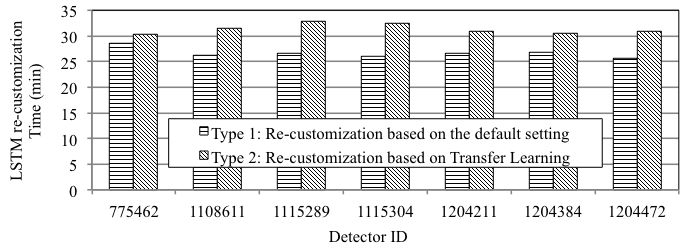}
  \caption{The required LSTM re-customization time for the seven detectors.}\label{fig:Figure12}
\end{figure*}
\begin{figure*}[!ht]
  \centering
  \includegraphics[width=0.6\textwidth]{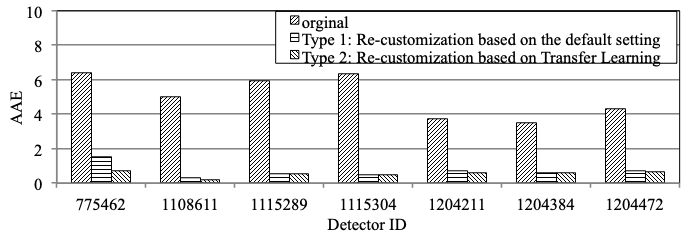}
  \caption{The AAE results before and after two types of LSTM re-customization are individually performed.}\label{fig:Figure13}
\end{figure*}
\begin{figure*}[!ht]
  \centering
  \includegraphics[width=0.6\textwidth]{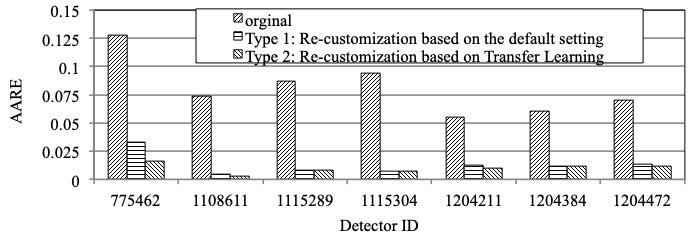}
  \caption{The AARE results before and after two types of LSTM re-customization are individually performed.}\label{fig:Figure14}
\end{figure*}
\begin{figure*}[!ht]
  \centering
  \includegraphics[width=0.6\textwidth]{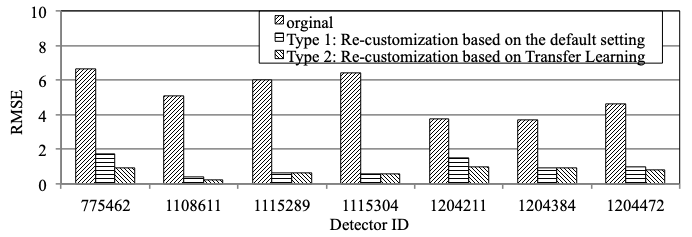}
  \caption{The RMSE results before and after two types of LSTM re-customization are individually performed.}\label{fig:Figure15}
\end{figure*}
Table 4 lists the hyperparameter settings separately used in Type 1 and Type 2 for the seven detectors. In fact, all the detectors inherited their LSTM models from the same detector (i.e., detector 1118333) since the traffic speed patterns they observed are similar to the one observed by detector 1118333. This is why these detectors have the same hyperparameter setting in Type 2.

Figure 12 shows the LSTM re-customization time for the seven detectors. Surprisingly, Type 2 consumes more time than Type 1 for every detector. The key reason is that the hyperparameter setting in Type 2 contains higher values for \emph{N\textsubscript{Unit}} and \emph{ep}, therefore prolonging the LSTM training time and increasing the required LSTM re-customization time for each detector.

Figures 13, 14, and 15 illustrate the prediction performance of the seven detectors before and after the two types of re-customization were individually performed. Both Type 1 and Type 2 are able to customize an appropriate LSTM model for each of these detectors and considerably reduce all the AAE, AARE, and RMSE values. It is clear that Type 2 leads to slightly better prediction accuracy than Type 1, but Type 2 consumes more processing time than Type 1. Therefore, choosing Type 1 seems to be more economic. For this reason, the re-customization approach of DistTune is based on Type 1, i.e., the default hyperparameter setting.

\section{Discussion}
In this section, we discuss and explain why we chose to use the traffic speed data of five continuous working days to be the training dataset for every detector in all our experiments, rather than using a longer period of data. According to our observation, not all detectors observe similar traffic speed pattern all the time. To demonstrate this point, we study one detector, Detector 718086, from freeway I5-N as an example. Figures 16, 17, 18, and 19 illustrate the traffic speed data collected by this detector on freeway I5-N between 4 a.m. and 10 a.m. for 1-week working days, 4-week working days, 8-week working days, and 12-week working days, respectively. When the observation period is one week, we can see from Figure 16 that the traffic speed patterns of these five working days have some deviation. This situation gets worse when we increased the observation time. Please see Figures 17, 18, and 19. It is clear that the traffic speed pattern collected by this detector has more variation as the observation period prolongs. Note that this phenomenon not only appear for this detector, we found that it also happens for many other detectors.

To show the effect of different lengths of training data, we evaluate the corresponding LSTM training time and prediction accuracy in terms of AAE, AARE, and RMSE based on the default hyperparameter setting (i.e., \emph{R\textsubscript{Learn}}=0.01, \emph{N\textsubscript{Layer}}=1, \emph{N\textsubscript{Unit}}=2, \emph{ep}=100). As shown in Figure 20, the 1-week scenario requires the shortest LSTM training time (around 81.8 seconds) because the size of the training data is only five days, which is the shortest among all the four scenarios. The time is only 28.4\%($=\frac{81.8}{228}$), 11.5\%($=\frac{81.8}{712}$), and 9.2\%($=\frac{81.8}{893}$) compared with the training time required in the 4-week scenario, 8-week scenario, and 12-week scenario, respectively.

On the other hand, from Figure 21, we can also see that the 1-week scenario outperforms the other three scenarios when it comes to prediction accuracy. Apparently, the 1-week scenario leads to the smallest value in AAE, AARE, and RMSE, implying that it leads to the best prediction accuracy among the four scenarios. 
\begin{figure}[!ht]
  \centering
  \includegraphics[width=0.48\textwidth]{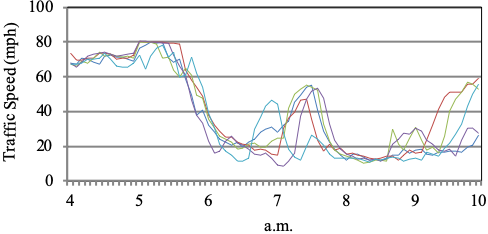}
  \caption{The traffic speed pattern of Detector 718086 between 4 a.m. and 10 a.m. for 1-week consecutive working days (from Oct 16, 2017 to Oct. 20, 2017).}\label{fig:Figure16}
\end{figure}
\begin{figure}[!ht]
  \centering
  \includegraphics[width=0.48\textwidth]{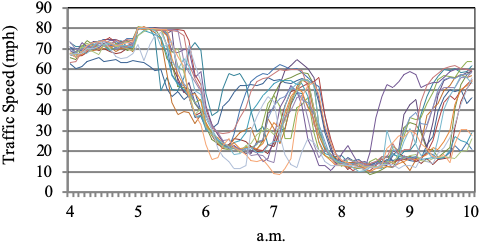}
  \caption{The traffic speed pattern of Detector 718086 between 4 a.m. and 10 a.m. for 4-week consecutive working days (from Sept. 25, 2017 to Oct. 20, 2017, i.e., the 5 working days of 4 consecutive weeks).}\label{fig:Figure17}
\end{figure}
\begin{figure}[!ht]
  \centering
  \includegraphics[width=0.48\textwidth]{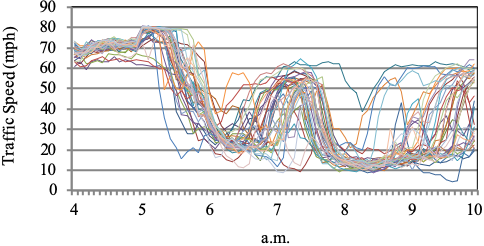}
  \caption{The traffic speed pattern of Detector 718086 between 4 a.m. and 10 a.m. for 8-week consecutive working days (from Aug. 28, 2017 to Oct. 20, 2017, i.e., the 5 working days of 8 consecutive weeks).}\label{fig:Figure18}
\end{figure}
\begin{figure}[!ht]
  \centering
  \includegraphics[width=0.48\textwidth]{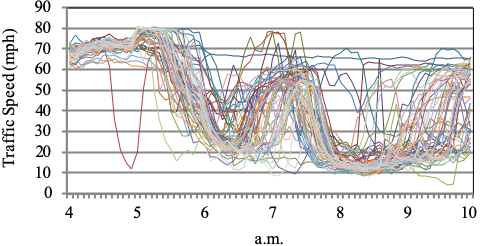}
  \caption{The traffic speed pattern of Detector 718086 between 4 a.m. and 10 a.m. for 12-week consecutive working days (from July 31, 2017 to Oct. 20, 2017, i.e., the 5 working days of 12 consecutive weeks).}\label{fig:Figure19}
\end{figure}
\begin{figure}[!ht]
  \centering
  \includegraphics[width=0.48\textwidth]{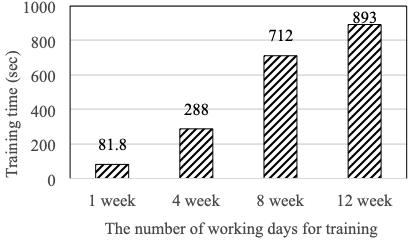}
  \caption{The required training time given different lengths of training data.}\label{fig:Figure20}
\end{figure}
\begin{figure*}[!ht]
  \centering
  \includegraphics[width=0.6\textwidth]{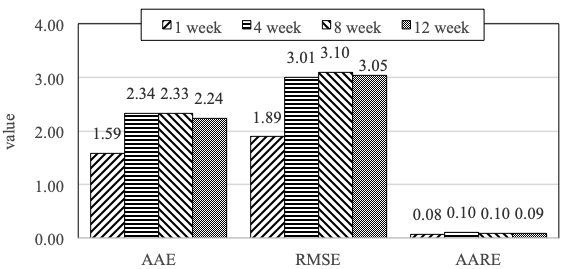}
  \caption{The prediction accuracy under different lengths of training data.}\label{fig:Figure21}
\end{figure*}

Based on the above results, we confirm that choosing 1-week of traffic speed data to be the training-data size of LSTM not only saves time, but also achieves higher prediction accuracy because of less deviation in the traffic speed patterns. These two properties are essential since DistTune is designed to provide time-efficient and accurate traffic speed prediction. 
\section{CONCLUSION AND FUTURE WORK}
In this paper, we have introduced DistTune, a distributed scheme to achieve fine-grained, accurate, time-efficient, and adaptive traffic speed prediction for the increasing number of detectors deployed in a growing transportation network. DistTune automatically customizes LSTM models for detectors based on NMM, which was chosen over BOA based on our empirical comparison and evaluation. By allowing LSTM models to be shared between different detectors and employing parallel processing, DistTune successfully addresses the scalability issue, and enables fine-grained and efficient traffic speed prediction. Furthermore, DistTune keeps monitoring detectors and re-customizes their LSTM models when necessary to make sure that their prediction accuracy can be always achieved and guaranteed. Our extensive experiments based on traffic data collected by the Caltrans performance measurement system demonstrate the performance of DistTune and confirm that DistTune provides fine-grained, accurate, time-efficient, and adaptive traffic speed prediction for a growing transportation network. 

In this paper, DistTune is implemented on a cluster consisting of only one master server and a set of worker nodes. Such a cluster might have some risks or limitations as follows. The master server is a single point of failure (SPOF), and it might crash or fail due to diverse reasons. In addition, its computation resources might not be able to support the operation of DistTune if DistTune is employed in a large-scale transportation network. On the other hand, the number of worker nodes decides the execution performance of DistTune. In this paper, the number of worker nodes is fixed for simplicity. However, it should be dynamically adjusted over time according to the workload of DistTune.

Therefore, in our future work, we would like to further address the SPOF issue and suggest a highly scalable and elastic solution based on a cloud or  a multi-cloud environment for DistTune. Furthermore, we plan to improve the performance of DistTune by considering proper scheduling approaches, such as \cite{26,27}. In addition, we plan to investigate Early Stopping \cite{28} and study its impact on the performance of DistTune. Furthermore, we would like to integrate DistTune with other novel techniques such as vehicle trajectory extraction \cite{38} to provide better traffic managements and services for growing transportation networks.

\begin{acks}
This work was supported by the eX\textsuperscript{3} project - \emph{Experimental Infrastructure for Exploration of Exascale Computing}, funded by the Research Council of Norway under contract 270053, and the scholarship under project number 80430060 supported by Norwegian University of Science and Technology. The authors also want to thank the anonymous reviewers for their reviews and valuable suggestions to this paper.
\end{acks}

\section{Author Contribution Statement}
The authors confirm contribution to the paper as follows: study conception and design: Ming-Chang Lee and Jia-Chun Lin; data collection: Ming-Chang Lee; analysis and interpretation of results: Ming-Chang Lee and Jia-Chun Lin; draft manuscript preparation: Ming-Chang Lee, Jia-Chun Lin, and Ernst Gunnar Gran. All authors reviewed the results and approved the final version of the manuscript.

\end{document}